# Double Machine Learning for Adaptive Causal Representation in High-Dimensional Data

Lynda Aouar [ab1] , Han Yu [b1]
[a] Yale University ,   [b] University of Northern Colorado

**ABSTARCT**

Adaptive causal representation learning from observational data is presented, integrated with an efficient sample splitting technique within the semiparametric estimating equation framework. The support points sample splitting (SPSS), a subsampling method based on energy distance, is employed for efficient double machine learning (DML) in causal inference. The support points are selected and split as optimal representative points of the full raw data in a random sample, in contrast to the traditional random splitting, and providing an optimal sub-representation of the underlying data generating distribution. They offer the best representation of a full big dataset, whereas the unit structural information of the underlying distribution via the traditional random data splitting is most likely not preserved. Three machine learning estimators were adopted for causal inference, support vector machine (SVM), deep learning (DL), and a hybrid super learner (SL) with deep learning (SDL), using SPSS. A comparative study is conducted between the proposed SVM, DL, and SDL representations using SPSS, and the benchmark results from Chernozhukov et al. (2018), which employed random forest, neural network, and regression trees with a random k-fold cross-fitting technique on the 401(k)-pension plan real data. The simulations show that DL with SPSS and the hybrid methods of DL and SL with SPSS outperform SVM with SPSS in terms of computational efficiency and the estimation quality, respectively.

## 1. Introduction

Economists, statisticians, and social scientists have developed models to estimate the effect of the target policy parameter in observational study. Firpo (2007) introduced a double-staged method to estimate the quantile treatment effect, which is based on estimating initially the nuisance parameter and then the estimation of the quantile treatment effect of interest. This method has shown a limitation in the cases when there is a high dimensional confounder, in which the sample size is much smaller than the number of the nuisance parameters ($p >> N$).

Chernozhukov et al. (2018) proposed a double/debiased machine learning (DML) method as an extension to Firpo's work (2007) based on the work of Belloni et al. (2012), Belloni et al. (2014), Chernozhukov et al. (2015), and Belloni et al. (2017). DML is a two-step causal inference method using observational data to estimate the average treatment effect (ATE). A two-staged bias correction is carried out by using the Neyman orthogonalization and moment score function to address the regularization bias of the target estimator (Klosin, 2021, Chernozhukov et al., 2022). Sample splitting as a cross-fitting technique overcomes the bias introduced by the model overfitting dilemma (Bach et al., 2022). The Neyman orthogonality delivers an estimation of $\sqrt{n}$ rate of convergence to the target parameter and allows an asymptotic normality (Lewis & Syrgkanis, 2021).

---

[1] Corresponding authors.
lynda.aouar@yale.edu (L Aouar). han.yu@unco.edu (H Yu).

This study is focused on statistical adaptive learning (AL), to a scale of regularity classes in semiparametric framework (Bickel et al., 2000; Van der Laan et al., 2004; Chambaz et al., 2016). The SPSS method is implemented as it is an optimal adaptation to the data distribution versus the random splitting (Mak & Joseph, 2018). DML is considered for causal inference, and Figure 1 summarizes the study's workflow diagram as follows,

**Figure 1**
*Study Work Map*

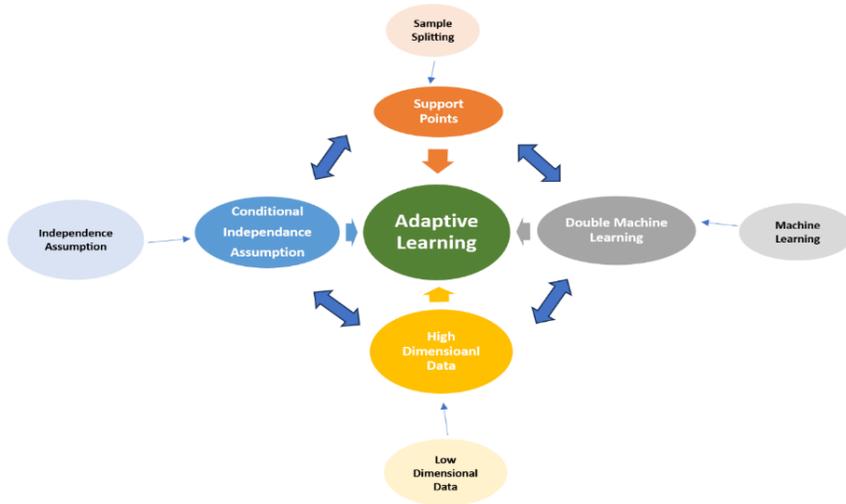

## Structural Causal Models

The structural causal model is represented as follows,

$$Y = T\beta_0 + g_0(\mathbf{X}) + U, \qquad \mathbb{E}[\,U \mid \mathbf{X}, T\,] = 0,$$

$$T = m_0(\mathbf{X}) + V, \qquad \mathbb{E}[\,V \mid \mathbf{X}\,] = 0,$$

where Y is the outcome variable, $\mathbf{X} = (X_1, ..., X_p)$ is the covariate vector, and T is the treatment. $\mathbf{X}$ as confounder affects the outcome Y as well as the treatment $T$ through the functions $g_0(.)$ and $m_0(.)$, respectively. $\eta_0 = (g_0(.), m_0(.))$ is an infinite dimensional nuisance parameter. $\beta_0$ is the parameter for the causal effect of $T$. $U$ and $V$ are the disturbances.

Chernozhukov et al. (2018) proposed the DML method for the structural causal model under the following assumptions,

1. $\beta_0$ satisfies the moment conditions as follows

$$E_P[\psi(Z; \beta_0, \eta_0)] = 0,$$

where $\eta_0$ is the true value of the nuisance parameter $\eta \in \mathcal{N}$, $\beta_0$ *is* the true value of the causal parameter $\beta \in \Theta \subseteq \mathbb{R}^{d_\beta}$, and $\Theta$ is non-empty measurable set.

2. $(Z_i)_{i=1}^N$ is a sequence of random observational data from the distribution of $Z$, where Z is a random element in the measurable space $(\mathcal{Z}, \mathcal{A}_Z)$ with probability measure $P \in \mathcal{P}_N$.
3. The vector of known score functions $\psi = (\psi_1, \ldots, \psi_{d_\beta})'$ such that $\psi_j$, j=1, ..., $d_\beta$, are functions defined on $\mathcal{Z} \times \Theta \times \mathcal{N}$ and measurable if assigning $\Theta$ and $\mathcal{N}$ with their Borel $\sigma$-fields.

***Definition 1.*** Consider $(I_k)_{k=1}^K$, a K-fold random partition of a sample of **N cases. Let** the complement set $I_k^c$ for each $I_k$: $I_k^c = \{1, \ldots, N\} \setminus I_k$, k$\in\{1,\ldots,$ **K**$\}$. Each $I_k$ has a size of n = N/K.

1. Construct the machine learning estimator of $\eta_0$,

$$\hat{\eta}_{0,k} = \hat{\eta}_0\left((Z_i)_{i \in I_k^c}\right), k \in \{1, \ldots, K\}.$$

2. Take $E_{n,k}[\psi(Z)] = n^{-1}\sum_{i \in I_k}\psi(Z_i)$ as the empirical expectation of the $k^{th}$ fold, where $\psi$ is the Neyman orthogonal score function, calculate the $k^{th}$ target parameter estimator $\check{\beta}_{0,k}$ that satisfies,

$$\mathbb{E}_{n,k}[\psi(Z; \check{\beta}_{0,k}, \hat{\eta}_{0,k})] = 0.$$

3. Construct the target estimator that is a combination of the k estimators, called the DML1 estimator, as follows,

$$\tilde{\beta}_0 = \frac{1}{K}\sum_{k=1}^K \check{\beta}_{0,k}.$$

4. Alternatively, in step 2, the target estimator, called the DML2 estimator, is the solution of the following equation:

$$\frac{1}{K}\sum_{k=1}^K E_{n,k}[\psi(Z; \tilde{\beta}_0, \hat{\eta}_{0,k})] = 0.$$

## 2. Related Work

Since the seminal work of Chernozhukov et al. (2018), many studies have emerged. Guo et al. (2022) developed doubly debiased Lasso based on observational data in the presence of high-dimensional unobserved covariates. They corrected the bias arising from both the high dimensional and hidden nuisance variables. Ju et al. (2018) applied various kinds of deep neural networks assigned with different layers of depth for each learner in the SL-method. A combined algorithm has been developed from deep neural networks and super learner methods (Young et al., 2018). Yang et al. (2020) studied double machine learning with SVM and k-fold cross-validation. A research study has been conducted by Kebonye (2021) using a combination of the two methods, SPSS and SVM. Varaku (2021) applied the mixture of DML with deep neural networks. A causal effect framework (Heiler & Knaus, 2022) applied double machine learning, deep learning, and k-fold cross-validation. DML is combined with SPSS for efficiency (Agboola & Yu, 2023). Alanazi (2022) used DML integrated with the super learner for adaptiveness. None of the previous studies have addressed the three methods proposed in this paper: the DML method using SVM, DL, and SDL, along with the hybrid SL with DL using SPSS, all together in a single study.

SVM and DL have been chosen in this study because they are known for their effectiveness and adaptiveness in tuning the hyperparameter. The hybrid methods of double machine learning and causal inference are nowadays a cutting-edge area in the practice and methodological studies (Knaus, 2021). This study aims to develop support points-based DML and to compare the proposed frameworks for estimating the average treatment effect (ATE) in structural causal models with the original DML approach introduced by Chernozhukov et al. (2018). The performance of the three methods, SVM, DL, and the hybrid SDL with the SPSS method, is compared with the k-fold sample splitting approach employed in Chernozhukov et al. (2018) to address the three research questions: How do the three methods perform on simulated data: DML using SVM with SPSS, DML using DL with SPSS, and DML using the hybrid SDL with SPSS. Real-world data will be used to demonstrate the three research questions, comparing the performance of each of the proposed methods with that of Chernozhukov, et al. (2018).

## Causation Inference

The causal relationship can be effectively deduced based on controlled randomized trials (CRT). However, in most cases, CRTs are not feasible due to unethical concerns, financial costs, or time constraints. Consequently, there is growing interest in learning causal relationships from observational data. This shift presents challenges in defining causal effects when cases are not randomly assigned to treatment or when control and treatment groups are absent. To address these issues, models such as the Structural Causal Model (SCM) and the Neyman-Rubin Potential Outcome framework have been developed. This study focuses on the SCM.

## Double Machine Learning

The key features of DML are outlined based on the theoretical foundations and proofs established by Belloni et al. (2017), Chernozhukov et al. (2018), and Bach et al. (2021). These features include the construction of the confidence regions for DML estimators, the variance estimator for causal target parameter for DML, semiparametric efficiency, the uniformly valid confidence interval of the scaler parameter for DML, and the inference for the partially linear regression model with DML.

Two theorems are presented to support the construction of confidence regions for the DML, DML1 or DML2.

*Assumption 1.* Approximate Neyman Orthogonality and Linear Scores: Suppose the score functions are linear as follows,

$$\psi(z;\beta,\eta) = \psi^a(z;\eta)\beta + \psi^b(z;\eta), \text{ for all } z \in \mathcal{Z}, \beta \in \Theta, \eta \in T.$$

Let $\{\mathcal{P}_N\}_{N\geq 1}$ a sequence of sets of distributions $P$ on $\mathcal{Z}$. Consider $\{\Delta_N\}_{N\geq 1}$ and $\{\delta_N\}_{N\geq 1}$ be two convergent sequences that tend to zero and are both sequences of positive constant. The constants $c_0, c_1, s, K$ ( fold size), and q are positive, where $c_0 \leq c_1$, $K \geq 2$, $q > 2$. Then $\forall N \geq 3, \forall P \in \mathcal{P}_N$ the following hold,

1. The true parameter $\beta_0$ satisfies

$$E_P \psi(W;\beta_0,\eta_0)[\eta - \eta_0] = 0.$$

2. The matrix $J_0 := E_P[\psi^a(Z;\eta_0)]$ has singular values $\in [c_0, c_1]$, which means that the identification condition holds.

3. The function $E_P[\psi(W;\theta,\eta)]$ with respect to $\eta$ is twice continuously Gateaux-differentiable on $T$.

4. The score function $\psi$ holds for the Neyman orthogonality. Or the score function $\psi$ obeys at $(\beta_0,\eta_0)$ the Neyman near-orthogonality condition $\lambda_N$ with respect to $\eta$ such that,

$$\lambda_N := \sup_{\eta \in \mathcal{T}_N} \|\partial_\eta E_P \psi(Z;\beta_0,\eta_0)[\eta - \eta_0]\| \leq \delta_N N^{-1/2}.$$

*Assumption 2.* The Quality of the Nuisance Parameter Estimator and the Score Regularity. Suppose a random fold $I \subset [N] = \{1,\ldots,N\}$ **of** size $\boldsymbol{n = N/K}$, $\forall\, N \geq 3$, $\forall\, P \in \mathcal{P}_N$, the following hold

1. The eigenvalues of the following matrix $\boldsymbol{E_P[\psi(Z;\beta_0,\eta_0)\psi(Z;\beta_0,\eta_0)']}$ are bounded from below by $c_0$. In other words, the score function $\boldsymbol{\psi}$ has a non-degenerate variance.

2. Given $\mathcal{T}_N$ the realization set of the nuisance parameter $\hat{\boldsymbol{\eta}}_{0,k} = \hat{\boldsymbol{\eta}}_0\left((Z_i)_{i \in I^c}\right)$, then

$$P[\,\hat{\eta}_{0,k} = \hat{\eta}_0\left((Z_i)_{i \in I^c}\right) \in \mathcal{T}_N) \geq 1 - \Delta_N.$$

3. $\mathcal{T}_N$ contains $\boldsymbol{\eta_0}$ and satisfies the moment conditions,

$$m_N := \sup_{\eta \in \mathcal{T}_N} \left(E_P[\|\psi(Z;\beta_0,\eta)\|^q]\right)^{1/q} \leq c_1,$$

$$m'_N := \sup_{\eta \in \mathcal{T}_N} \left(E_P[\|\psi^a(Z;\eta)\|^q]\right)^{1/q} \leq c_1$$

Also, the next inequalities specify the statistical rates $\lambda'_N$, $r_N$, $r'_N$, respectively,

$$\lambda'_N := \sup_{r \in (0,1), \eta \in \mathcal{T}_N} \|\partial_r^2 E_P[\psi(Z;\beta_0,\eta_0 + r(\eta - \eta_0))]\| \leq \delta_N/\sqrt{N},$$

$$r_N := \sup_{\eta \in \mathcal{T}_N} \|E_P[\psi^a(Z;\eta)] - E_P[\psi^a(Z;\eta_0)]\| \leq \delta_N,$$

$$r_N' := \sup_{\eta \in \mathcal{T}_N} \left(E_P[\|\psi(Z;\beta_0,\eta) - \psi(Z;\beta_0,\eta_0)\|^2]\right)^{1/2} \leq \delta_N,$$

which means that under the assumption 2, and for a chosen value of $\varepsilon_N$ such that $\|\hat{\eta}_0 - \eta\|_T \lesssim \varepsilon_N$ in the realization set $\mathcal{T}_N$, where the function $\psi: (\beta,\eta) \mapsto \psi(Z;\beta,\eta)$ is smooth, by taking $\lambda'_N \lesssim \varepsilon_N^2$, $r_N \lesssim \varepsilon_N$, $r'_N \lesssim \varepsilon_N$, then when considering a special case where $\lambda'_N = o(N^{-1/2})$ it will follow that $\varepsilon_N = o(N^{-1/4})$. Thus, the nuisance parameter estimator $\hat{\eta}_0$ has the $N^{-1/4}$ rate of convergence.

**Remark.** Assumption 1 is required to make sure that the score functions are Neyman orthogonal or approximately orthogonal, and mild smoothness. Assumption 2 is about the quality of the nuisance parameter estimator and the score function regularity condition. The first theorem addresses the asymptotic normality of the estimator.

**Theorem 1.** Under assumption 1 and assumption 2, $\forall\ N$, let $\delta_N \geq 1/\sqrt{N}$. The DML1 estimator $\tilde{\beta}_0$ (and the DML2) has the asymptotic normality distribution property with a $\sqrt{N}$ convergence,

$$\sqrt{N}\sigma^{-1}(\tilde{\beta}_0 - \beta_0) = \frac{1}{\sqrt{N}}\sum_{i=1}^{N} \bar{\psi}(Z_i) + O_P(\rho_N) \xrightarrow{d} N(0, I_d),$$

where the approximate variance and the influence function are respectively

$$\sigma^2 := J_0^{-1} E_P[\psi(Z; \beta_0, \eta_0)\psi(Z; \beta_0, \eta_0)'](J_0^{-1})',$$

$$\bar{\psi}(\cdot) := -\sigma^{-1} J_0^{-1} \psi(\cdot, \beta_0, \eta_0),$$

and the remainder $\rho_N$ satisfies

$$\rho_N := N^{-1/2} + r_N + r'_N + N^{1/2}\lambda_N + N^{1/2}\lambda'_N \lesssim \delta_N.$$

The theorem shows that the estimator, $\hat{\beta}_0$, based on the orthogonal scores, will reach a convergence of $\sqrt{N}$ rate and will have normal distribution asymptotically. This distributional approximation and concentration rate are both maintained uniformly in $\mathcal{P}_N$, where $\mathcal{P}_N$ is an expanding class of probability measures, $(P_N)_{N\geq 1}$ is a sequence of probability distributions such that for each N, $P_N \in \mathcal{P}_N$, and $P$ is varying over $\mathcal{P}_N$.

The second theorem identifies the variance estimator.

**Theorem 2.** Under the criteria of assumption 1 and assumption 2, $\forall\ N$, let $\delta_N \geq N^{-[(1-2/q)\wedge 1/2]}$. Then the asymptotic variance matrix of the $\sqrt{N}(\tilde{\beta}_0 - \beta_0)$ is

$$\hat{\sigma}^2 = \hat{J}_0^{-1} \frac{1}{K}\sum_{k=1}^{K} E_{n,k}\left[\psi(Z; \tilde{\beta}_0, \hat{\eta}_{0,k})\psi(Z; \hat{\beta}_0, \hat{\eta}_{0,k})'\right](\hat{J}_0^{-1})',$$

where

$$\hat{J}_0 = \frac{1}{K}\sum_{k=1}^{K} E_{n,k}[\psi^a(Z; \hat{\eta}_{0,k})].$$

And

$$\hat{\sigma}^2 = \sigma^2 + O_P(\varrho_N),$$

$$\varrho_N := N^{-[(1-2/q)\wedge 1/2]} + r_N + r'_N \lesssim \delta_N,$$

which allows to substitute $\sigma^2$ by $\hat{\sigma}^2$ with a remainder,

$$\rho_N = N^{-[(1-2/q)\wedge 1/2]} + r_N + r'_N + N^{1/2}\lambda_N + N^{1/2}\lambda'_N.$$

# Semiparametric Method

Semiparametric methods are the methods developed for a class of statistical models that have the parametric and the nonparametric components by adopting assumptions that fully define the distribution. However, semiparametric models still require minimum structure (Max & Zang, 2019). Specifically, a statistical model is a class of probability measures $\{P, \ P \in \mathcal{P}\}$ on a sample space $\chi$. Assume that $\mathbf{P}$ is indexed by a parameter space $\Theta$, for each $\theta \in \Theta$, $P_\theta$ is specified such that $\mathbf{P} = \{P_\theta, \ \theta \in \Theta\}$. Thus, the statistical model $\mathbf{P}$, which is indexed by $\theta \in \Theta$, is considered parametric if $\Theta \subseteq \mathbb{R}^k$, the Euclidean space of k-dimensional for a positive integer k(Bickel et al., 2006). And it is a nonparametric model if the space of the parameters $\Theta \subseteq \mathbf{H}$, where $\mathbf{H}$ is an infinite-dimensional space. The statistical models are defined as semiparametric models $\{P_{\theta,\eta} : \theta \in \Theta, \ \eta \in \mathbf{H}\}$ if they have one or more finite-dimensional parameter constituents $\theta \in \Theta$, and one or more infinite-dimensional parameter elements $\eta \in \mathbf{H}$, where $\mathbf{H}$ is a space of functions. $\theta \in \Theta \subseteq \mathbb{R}^k$ is the parameter of interest, and $\eta \in \mathbf{H}$ is the infinite-dimensional nuisance parameter (Bickel et al., 2000; Kosorok, 2006). For instance, assume the semiparametric regression model $Y = \beta Z + \epsilon$, where $\beta$ is the k-dimensional Euclidean space parameter defining the parametric statistical components in the model (Kosorok, 2006).

*Corollary 1.* In general conditions, the semiparametric efficiency of the target estimator is not required, however, in special cases it could attain the semiparametric efficiency. If the condition of theorem 1 is met, and if the score function $\psi$ is efficient for $\tilde{\beta}_0$ under the semiparametric paradigm (Van der Laan & McKeague,1998) at specific $P \in \mathcal{P} \subset \mathcal{P}_N$, then the variance $\sigma_0^2$ of $\tilde{\beta}_0$ attains the bounds of the semiparametric efficiency at $P$ relative to $\mathcal{P}$.

*Corollary 2.* Uniformly Valid Confidence Interval of Scalar Parameter estimator of DML: If the conditions of theorem 2 hold, then, for some vector $\ell_{d_\beta \times 1}$, the constructed confidence interval for the scalar parameter $\ell'\beta_0$ will be as follows

$$\text{CI} := \left(\ell'\tilde{\beta}_0 \pm \Phi^{-1}(1 - \alpha/2)\sqrt{\ell'\hat{\sigma}^2\ell/N}\right),$$

that satisfies

$$\sup_{P \in \mathcal{P}_N} |\Pr_P (\ell'\beta_0 \in \text{CI}) - (1 - \alpha)| \to 0,$$

which means that $\forall \ P_N \in \mathcal{P}_N$, the confidence interval obeys,

$$\Pr_{P_N} (\ell'\beta_0 \in \text{CI}) \to (1 - \alpha).$$

Thus, the confidence interval is uniformly valid. If $\epsilon_N \to 0$,

$$\sup_{P \in \mathcal{P}_N} |\Pr_P (\ell'\beta_0 \in \text{CI}) - (1 - \alpha)| \leq \left|\Pr_{P_N} (\ell'\beta_0 \in \text{CI}) - (1 - \alpha)\right| + \epsilon_N \to 0.$$

# 3. Methodology

First, the support points subsampling is presented heuristically. Second, the models under study include SVM, DL, and an ensemble method that combines SL, DL, and SPSS. Third, the double machine learning for causal inference is described (Chernozhukov et al., 2018) with the sample splitting technique to develop the target estimator after the estimation of the nuisances.

## The Energy Distance

*Definition 2.* Assume $\mathbf{V} = (\mathbf{U}, Y)$ is a continuous random vector. The energy distance between the empirical distribution of points $\mathbf{v_1}, \mathbf{v_2}, \ldots, \mathbf{v_n}$ and the distribution $G(V)$ is described as follows,

$$ED = \frac{2}{n}\sum_{i=1}^{n} \mathbb{E}\|\mathbf{v}_i - \mathbf{V}\|_2 - \frac{1}{n^2}\sum_{i=1}^{n}\sum_{j=1}^{n} \|\mathbf{v}_i - \mathbf{v}_j\|_2 - \mathbb{E}\|\mathbf{V} - \mathbf{V}'\|_2 ,$$

where $\|.\|_2$ is the Euclidean distance. $\mathbf{V}, \mathbf{V}'$ are distributed as the distribution $G$. And all the expectation has been taken with respect to $G$ taking into consideration that all variables should be standardized with mean zero and the unit standard deviation to calculate the Euclidean distance. Mak and Joseph (2018) have noted that the energy distance will be small in the case that $\mathbf{v_1}, \mathbf{v_2}, \ldots, \mathbf{v_n}$ are close to $G$. So, they have expressed the minimizer of the energy distance to be the support points definition as follows,

$$\{\mathbf{v}_i^*\}_{i=1}^{n} \in \underset{\mathbf{v_1},\mathbf{v_2},\ldots,\mathbf{v_n}}{\operatorname{argmin}} ED = \underset{\mathbf{v_1},\mathbf{v_2},\ldots,\mathbf{v_n}}{\operatorname{argmin}} \left\{ \frac{2}{n}\sum_{i=1}^{n} \mathbb{E}\|\mathbf{v}_i - \mathbf{V}\|_2 - \frac{1}{n^2}\sum_{i=1}^{n}\sum_{j=1}^{n} \|\mathbf{v}_i - \mathbf{v}_j\|_2 \right\}.$$

## Support Points Sample Splitting

We begin by introducing sample splitting with support points. The following definitions and assumptions, adapted from Joseph & Vakayil (2021) and Székely & Rizzo (2013), are tailored to this study.

Suppose that a sample unit data structure $\mathbf{S} = \{(\mathbf{U}_i, Y_i)\}_{i=1}^{N}$ that consists of the predictor $\mathbf{U} = (\mathbf{T}, X)$ of dimension p, where T is the treatment, and $Y$ response. The aim is to perform the sample splitting with the support points method and divide the data into two mutually exclusive and disjoint sets of $\mathbf{S}$, a training set $\mathbf{S_1}$ and test set $\mathbf{S_2}$ such that

$$N = \mathbf{n}_{train} + \mathbf{n}_{test} = \operatorname{card}(\mathbf{S_1}) + \operatorname{card}(\mathbf{S_2}), \ \mathbf{S} = \mathbf{S_1} \cup \mathbf{S_2} \text{ and } (\mathbf{S_1})^c = \mathbf{S_2}.$$

Assume that the samples are independent and identically distributed from a distribution $\mathbf{G}$,

$$(\mathbf{U}_i, Y_i) \overset{iid}{\sim} G, i = 1, \ldots, N.$$

Let $H(\mathbf{U}; \boldsymbol{\theta})$ be the adaptive predictor from the dataset, $\boldsymbol{\theta}$ is the parameter vector to be estimated from the loss function $L(Y, H(\mathbf{U}; \boldsymbol{\theta}))$. The loss function can be taken as the squared or absolute error loss, or the negative predictor log-likelihood. The wish is that the adaptive predictor $H(\mathbf{U}; \boldsymbol{\theta})$ is near to the true predictor E (Y | U) under some specific $\boldsymbol{\theta}$. So, take the training sample to train multiple predictive models and then test their performance. The unknown vector parameter could be estimated by

$$\widehat{\boldsymbol{\theta}} = \underset{\boldsymbol{\theta}}{\operatorname{argmin}} \frac{1}{n_{\text{train}}} \sum_{i=1}^{n_{\text{train}}} L\left(Y_i^{\text{train}}, H(\mathbf{U}_i^{\text{train}}; \boldsymbol{\theta})\right)$$

given the training dataset,

$$\left(\mathbf{U}_i^{\text{train}}, Y_i^{\text{train}}\right) \sim G, i = 1, \dots, n_{\text{train}}.$$

The performance of the models could be evaluated by calculating the generalization error (Hastie et al., 2009),

$$\mathcal{E} = E_{\mathbf{U},Y}\{L(Y, H(\mathbf{U}; \widehat{\boldsymbol{\theta}})) \mid \mathcal{S}^{\text{train}}\}.$$

And given that the testing dataset is from,

$$(\mathbf{U}_i^{\text{test}}, Y_i^{\text{test}}) \sim G, i = 1, \dots, n_{\text{test}},$$

estimate this error from the testing set $\mathcal{S}^{\text{test}}$,

$$\widehat{\mathcal{E}} = \frac{1}{n_{\text{test}}} \sum_{i=1}^{n_{\text{test}}} L\left(Y_i^{\text{test}}, H(\mathbf{U}_i^{\text{test}}; \widehat{\boldsymbol{\theta}})\right).$$

Thus, the estimation $\widehat{\mathcal{E}}$ will be a Monte Carlo (MC) estimator which decreases at a rate of $\mathcal{O}(1/\sqrt{N_{test}})$. However, Mak and Joseph (2018) introduced the support points method for sample splitting with a Quasi-Monte Carlo (QMC) sample. This method could improve the estimation of $\mathcal{E}$ with a faster convergence rate of $\mathcal{O}(1/N_{\text{test}})$. Furthermore, it could be applied on a sample from a general distribution not only limited to the uniform distribution (Niederreiter, 1992).

First, one of the features of the support points is that the expectation in the support points equation could be substituted with the Monte Carlo average that is computed from $S = \{(\boldsymbol{U}_i, Y_i)\}_{i=1}^{N}$, the sample set of interest (Joseph & Vakayil, 2021). This empirical substitution is designed to solve the difficulty of not having the exact distribution of $G$, which makes the support points a flexible data adaptive technique. Thus

$$\{\mathbf{v}_i^*\}_{i=1}^n \in \underset{\mathbf{v}_1,\mathbf{v}_2,\dots,\mathbf{v}_n}{\operatorname{argmin}} \left\{ \frac{2}{nN} \sum_{i=1}^n \sum_{j=1}^N \|\mathbf{v}_i - \mathbf{V}_j\|_2 - \frac{1}{n^2} \sum_{i=1}^n \sum_{j=1}^n \|\mathbf{v}_i - \mathbf{v}_j\|_2 \right\}.$$

Second, the support points method is regarded as the best n points set that could represent the data distribution $G$ based on the energy distance criteria (Mak & Joseph, 2018). It outperforms the other points splitting techniques such as the principal points method defined by Flury (1990), and *MSE*-rep method introduced by Fang and Wang (1994). Precisely, the support points converging in distribution to $G$, makes it as a QCM sample for $G$, which the two other methods do not have.

SPSS is an optimal adaptive subsample of a dataset that is well-representative of the underlying data-generating distribution ( Mak & Joseph, 2018). We outline the main theoretical features of support points within the measure-theoretic framework, defined on a probability measure space ( $\Omega, \mathcal{F}, P$) (Billinsley, 2012), as follows.

***Lemma 1.*** Consider a sequence of random variables $\mathbf{X_n} := \{X_j\}_{j=1}^n$ and a sequence of support points $\mathbf{S_n} := \{S_i\}_{i=1}^n$ with the empirical distribution function $G_n$ such that

$$\mathbf{X_n} \sim G_n,$$

$$\mathbf{S_n} \sim G_n,$$

$$\mathbf{X} \sim G.$$

Suppose $\varphi_n(t)$ and $\varphi(t)$ are the characteristic functions of $\mathbf{S_n}$ and $\mathbf{X}$ respectively. If

$$\lim_{n \to \infty} \varphi_n(t) = \varphi(t),$$

then

$$\mathbf{S_n} \xrightarrow{d} \mathbf{X}.$$

**Remark.** This lemma could be verified using Halley's theorem and Cramér-Lévy theorem.

***Proposition 1.*** The Existence of a Convergent Subsequence: Consider $\mathbf{X_n} := \{X_j\}_{j=1}^n$ a sequence of random variables where $\mathbf{X_n} \xrightarrow{d} \mathbf{X}$. Then there exists a subsequence $\mathbf{S}_{n_k}$ of support points $\mathbf{S_n} := \{S_i\}_{i=1}^n$ such that $\mathbf{S}_{n_k} \xrightarrow{d} \mathbf{X}$, which means that $\varphi_{n_k}(t) \to \varphi(t)$, and one of these subsequences is the support points subsequence.

**Remark.** Halley's theorem could be used to certify the existence of such subsequences. The following theorem 3, theorem 4, theorem 5, and theorem 6 show that one of these subsequences is the support points subsample that satisfies the conditions of this proposition.

***Theorem 3.*** Assume $\mathbf{S_n} := \{S_i\}_{i=1}^n$ is a sequence of independent and identically distributed (iid) support points-based random variables. *Let $\breve{G}_n$ and $G$* be the empirical distribution function (EDF) and limiting distribution function respectively with the corresponding characteristic functions $\breve{\varphi}_n(t)$ and $\varphi(t)$, then

$$\lim_{n \to \infty} \breve{\varphi}_n(t) = \varphi(t),$$

and

$$\lim_{n \to \infty} E[\,|\breve{\varphi}_n(t) - \varphi(t)|^2\,] = 0.$$

**Remark**. Given the EDF $\check{G}_n(x) = \frac{1}{n}\sum_{k=1}^{n} I(X_k \leq x)$, then by the Glinvenko-Cantelli Theorem,

$$\sup_x |\check{G}_n(x) - G(x)| \to 0 \text{ a.s.}$$

$$\Rightarrow \check{G}_n(x) \to G(x) \text{ a.s.}$$

By the Cramér- Lévy theorem, it can be deduced that,

$$\lim_{n\to\infty} \check{\varphi}_n(t) = \varphi(t), \text{ on any finite } |t| \in T \quad (*).$$

Also,

$$|e^{i\,xt}| \leq 1$$

$$\Rightarrow |\check{\varphi}_n(t)| \leq 1$$

$$\Rightarrow |\check{\varphi}_n(t) - \varphi(t)|^2 \leq c, \quad c \text{ is constant}.$$

As $|\check{\varphi}_n(t) - \varphi(t)|^2$ is bounded, then by the Portmanteau lemma the equation (*) will be as follows,

$$\lim_{n\to\infty} E\left[|\check{\varphi}_n(t) - \varphi(t))|^2\right] = 0.$$

***Theorem 4.*** Assume $\mathbf{S_n} := \{\mathbf{S_i}\}_{i=1}^n$ is a sequence of independent and identically distributed (iid) support points-based random variables, $\check{G}_n$ and $G$ are the empirical distribution function and its limiting distribution function respectively with the corresponding characteristic functions $\check{\varphi}_n(t)$ and $\varphi(t)$. Let $E_d(\check{G}_n, G)$ be the energy distance. Thus, the following holds,

$$\lim_{n\to\infty} E\left[E_d(\check{G}_n, G)\right] = 0,$$

where the definition of the energy distance is defined by Székely and Rizzo (2013),

$$E_d(\check{G}_n, G) = \frac{1}{K_p} \int \frac{|\check{\varphi}_n(t) - \varphi(t))|^2}{\|t\|_2^{p+1}} dt,$$

where

$$K_p = \frac{\pi^{p+1}}{\Gamma(\frac{p+1}{2})}.$$

**Remark.** The energy distance definition is as follows:

$$E_d(\check{G}_n, G) = \frac{1}{K_p} \int \frac{|\check{\varphi}_n(t) - \varphi(t))|^2}{\|t\|_2^{p+1}} dt,$$

where $E_d(\check{G}_n, G) < \infty$ (Székely & Rizzo, 2013)

$$\Rightarrow E\{E_d(\check{G}_n, G)\} = E\{\frac{1}{K_p}\int \frac{|\breve{\varphi}_n(t)-\varphi(t))|^2}{\|t\|_2^{p+1}}dt\}.$$

By Fubini theorem

$$E\{E_d(\check{G}_n, G)\} = \frac{1}{K_p}\int \frac{E[|\breve{\varphi}_n(t) - \varphi(t))|^2]}{\|t\|_2^{p+1}}dt.$$

which implies

$$\lim_{n\to\infty} E\{E_d(\check{G}_n, G)\} = \lim_{n\to\infty} \frac{1}{K_p}\int \frac{E[|\breve{\varphi}_n(t) - \varphi(t))|^2]}{\|t\|_2^{p+1}}dt$$

$$= \frac{1}{K_p}\int \lim_{n\to\infty} \frac{E[|\breve{\varphi}_n(t) - \varphi(t))|^2]}{\|t\|_2^{p+1}}dt,$$

by the dominated convergence theorem (DCT).
By theorem 3,

$$\lim_{n\to\infty} E[|\breve{\varphi}_n(t) - \varphi(t))|^2] = 0.$$

Thus,

$$\lim_{n\to\infty} E[E_d(\check{G}_n, G)] = 0.$$

***Theorem 5.*** Assume $S_n := \{S_i\}_{i=1}^n$ is a sequence of independent and identically distributed (iid) support points-based random variables, $G_n$, $\check{G}_n$, and $G$ are the cumulative distribution function (CDF), the empirical distribution function (EDF), and their limiting distribution function (DF) respectively. Consider the corresponding characteristic functions $\varphi_n(t)$, $\breve{\varphi}_n(t)$ and $\varphi(t)$, and their energy distance $E_d(G_n, G)$, the following holds,

$$\lim_{n\to\infty} \varphi_n(t) = \varphi(t).$$

**Remark.** From Mak and Joseph (2018), the energy distance has the following property:

$$0 \leq E_d(G_n, G) \leq E[E_d(\check{G}_n, G)].$$

And from theorem 4,
$$\lim_{n\to\infty} E[E_d(\check{G}_n, G)] = 0.$$

Thus,

$$0 \leq \lim_{n\to\infty} E_d(G_n, G) \leq \lim_{n\to\infty} E[E_d(\check{G}_n, G)] = 0$$

$$\Rightarrow \lim_{n\to\infty} E_d(G_n, G) = 0. \quad \ldots(i)$$

By the definition of the energy distance,

$$E_d(G_n, G) = \frac{1}{K_p} \int \frac{|\varphi_n(t) - \varphi(t)|^2}{\|t\|_2^{p+1}} dt$$

$$\Rightarrow \lim_{n \to \infty} E_d(G_n, G) = \lim_{n \to \infty} \frac{1}{K_p} \int \frac{|\varphi_n(t) - \varphi(t)|^2}{\|t\|_2^{p+1}} dt.$$

By the dominated convergence theorem, the following holds,

$$\lim_{n \to \infty} E_d(G_n, G) = \frac{1}{K_p} \int \lim_{n \to \infty} \frac{|\varphi_n(t) - \varphi(t)|^2}{\|t\|_2^{p+1}} dt. \quad (ii)$$

From (i) and (ii) that implies,

$$\lim_{n \to \infty} [|\varphi_n(t) - \varphi(t)|^2] = 0.$$

Then

$$\lim_{n \to \infty} \varphi_n(t) = \varphi(t).$$

***Theorem 6.*** Let the sequences $\mathbf{X_n} = \{X_j\}_{j=1}^n$, and $\mathbf{S_n} = \{S_i\}_{i=1}^n$ be the sample and the support points-based sample of random variables respectively, such that

$$\mathbf{X_n} \sim G_n,$$

$$\mathbf{X_n} \xrightarrow{d} \mathbf{X} \sim G,$$

$$\mathbf{S_n} \sim G_n.$$

Thus,

$$\mathbf{S_n} \xrightarrow{d} \mathbf{X} \sim G.$$

**Remark.** From proposition 1, theorem 3, theorem 4, and theorem 5, it can be concluded that the sequence of random variables $S_n$ of the support points satisfies,

$$\lim_{n \to \infty} \varphi_n(t) = \varphi(t).$$

By lemma 1,

$$\mathbf{S_n} \xrightarrow{d} \mathbf{X} \sim G.$$

*Corollary 3.* Let $\mathbf{X_n}$ and $\mathbf{S_n}$ be the sequences of the random sample and the support points sample respectively such that,

$$\mathbf{X_n} \sim G_n,$$

$$\mathbf{X_n} \xrightarrow{d} \mathbf{X} \sim G,$$

And

$$\mathbf{S_n} \sim G_n.$$

Suppose $f$ is continuous functions such that $f: (\Omega, \mathcal{F}, P) \to \mathbb{R}$, thus

$$f(\mathbf{S_n}) \xrightarrow{d} f(\mathbf{X}).$$

**Remark.** This corollary can be proved using theorem 4 and the continuous mapping theorem.

*Corollary 4.* Suppose the sequence $\mathbf{X_n}$ and $\mathbf{S_n}$ of random variables and the support points sample, respectively, such that

$$\mathbf{X_n} \sim G_n,$$

$$\mathbf{X_n} \xrightarrow{d} \mathbf{X} \sim G,$$

and

$$\mathbf{S_n} \sim G_n.$$

Suppose $f$ is continuous and bounded function such that $f: (\Omega, \mathcal{F}, P) \to \mathbb{R}$, thus

$$\lim_{n \to \infty} E[f(\mathbf{S_n})] = E[f(\mathbf{X})].$$

**Remark.** This corollary can be proved using the Portmanteau Lemma.

## Support Points Sample Splitting and Random Splitting

The application of the support points for splitting the dataset into training and test subsets has shown an optimal result versus the counterpart method of the random splitting (Joseph and Vakayil, 2021). Empirically, consider taking the training set larger than the test set, so, it will be more computationally efficient to create the test-set first. By implementing the equation stated earlier and taking $n = N_{test}$, thus,

$$\{\mathbf{v}_i^*\}_{i=1}^n \in \underset{\mathbf{v_1, v_2, \ldots, v_n}}{\mathrm{argmin}} \left\{ \frac{2}{nN} \sum_{i=1}^n \sum_{j=1}^N \|\mathbf{v}_i - \mathbf{V}_j\|_2 - \frac{1}{n^2} \sum_{i=1}^n \sum_{j=1}^n \|\mathbf{v}_i - \mathbf{v}_j\|_2 \right\}.$$

Figure 2 is the visualization of the test set of both support points sample splitting and random splitting, where we can observe that the support points splitting set is noticeably more representative of the original dataset than the random sample splitting set, which can deliver a much better estimation and inference accuracy (Székely & Rizzo, 2013) in learning and inference.

**Figure 2**
*Empirical Comparison Between the Random and Support Points-Based Splitting*

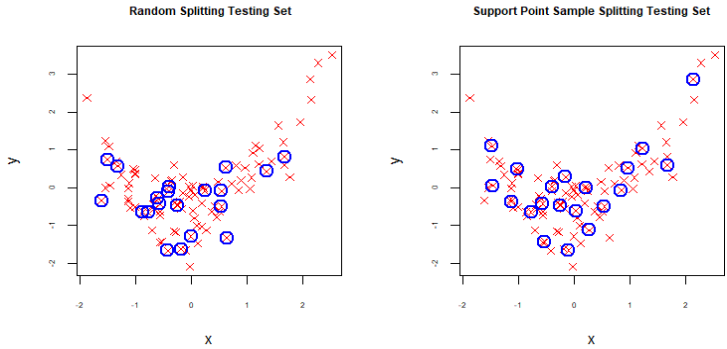

## A Hybrid of Super Learner and Deep Learning with Support Points

The deep neural network and super learner are designed for DML with SPSS, a method used as machine learning to estimate the causal target parameter in the SCM. Van der Laan et al. (2007) has extended the work of stacking from Wolpert (1992) to introduce what is called a super learner, which implements cross-validation and minimizes the validation risk to produce an optimal prediction based on an ensemble of learners which also has superior performance than to those learners individually.

Ensemble learning is a combination of multiple learners using specific procedures (Ju et al., 2018). Researchers in different fields have demonstrated increased interest in the ensemble methods due to their high performance in the prediction of empirical data, for example, the ensemble method application in an online learning study (Benkeser et al., 2018), mortality prediction study (Chambaz et al., 2016), and precision medicine study (Wyss et al., 2018; Alanazi, 2022). Typically, boosting, bagging, and stacking are examples of ensemble learning techniques. The boosting ensemble method takes care of the weak learner and boosts its performance (Freund & Schapire, 1996). Conversely, bagging ensemble methods take care of the strongest algorithm to minimize its variance by applying the bootstrap aggregation (Breiman, 1996). Stacking is the linear combination of all learners (Wolpert, 1992).

Super learner is an ensemble that estimates the performance of multiple algorithms through the cross-validation method, which has a result that is as good as the best-performing algorithm in the combination. It generates optimal weights for each learner in the ensemble based on their performance (Van Der Laan & Dudoit 2003; Van Der Laan et al., 2007). We summarize the super learner algorithm in the following steps,

1. Split data into $k$ blocks.
2. Fit all $M$ methods on blocks, leaving out one block.
3. On each block, calculate for each method the mean squared error (MSE).
4. Repeat $(k-1)$ times in steps 2 and 3.
5. leave out one block $j = 2,3,...,k$ for each repetition.
6. Choose the method with the lowest MSE through the blocks.

Each learner $L_k(n)$ ( k= 1,..., K(n))  from a collection of learners  $\widehat{\Psi}_k$  is an algorithm on the empirical distributions, i.e., a functional of the empirical distributions  $P_n$ in the parameter space $\Psi$,

$$L_k : P_n \to \widehat{\Psi}_k(P_n),$$

then, the super learner is defined:

$$\widehat{\Psi}_k(P_n) \equiv \widehat{\Psi}_{\hat{k}(P_n)}(P_n),$$

where $\hat{k}(P_n)$ is the selector that selects the optimal learner to minimize the cross-validation risk. Thus,

$$\hat{k}(P_n) \equiv \underset{k}{\operatorname{argmin}}\ E_{B_n} \sum_{i, B_n(i)=1} \left(Y_i - \widehat{\Psi}_k(P^0_{n,B_n})(X_i)\right)^2,$$

where $P^0_{n,B_n}$ is the empirical distribution of the validation set, $P^1_{n,B_n}$ is the empirical distribution of the training set, $B_n \in \{0,1\}^n$ is a random binary vector to define the split of validation and training learning. $\{i: B_n(i) = 0\}$ and $\{i: B_n(i) = 1\}$ are the validation and training samples.

Van der Laan et al. (2007) prove the following theorem that the super learner performs as best as the oracle selector up to the second order. So, the super learner is counted as the optimal learner when $L_k(n)$ learners are polynomial in the sample size (n) under the following assumptions.

***Assumption 3.*** The loss function $L(O, \psi) = (Y - \psi(X))^2$ should be uniformly bounded,

$$\sup_{\psi \in \Psi} \sup_O |L(O, \psi) - L(O, \psi_0)| \leq S_1,$$

where

$$\Psi(P_0) = \psi_0 \text{ is the true parameter.}$$

***Assumption 4.*** The variance of the $\psi_0$- centered loss function $L(O, \psi) - L(O, \psi_0)$ can be bounded by its expectation uniformly,

$$\sup_{\psi \in \Psi} \frac{\operatorname{var}_{P_0}|L(O,\psi) - L(O,\psi_0)|}{E_{P_0}|L(O,\psi) - L(O,\psi_0)|} \leq S_2,$$

***Theorem 7.*** Under assumption 3 and assumption 4, let p be the proportion of observations in the validation sample, specify $\{\widehat{\psi}_k = \widehat{\Psi}_k(P_n), k = 1, \ldots, K(n)\}$ as the set of $K(n)$ estimators, where the true parameter is defined as follows,

$$\psi_0 = \underset{\psi \in \Psi}{\operatorname{argmin}} \int L(o, \psi) dP_0(o).$$

The difference of risk between parameter $\psi_0$ and the candidate estimator $\psi$ is as follows,

$$d_0(\psi, \psi_0) \equiv E_{P_0}\{L(O, \psi) - L(O, \psi_0)\},$$

thus, for any $\lambda$, the expected risk difference is

$$\boldsymbol{E} d_0\left(\hat{\Psi}_{\hat{k}(P_n)}\left(P^0_{n,B_n}\right), \psi_0\right) \leq (1+2\lambda) E d_0\left(\hat{\Psi}_{\tilde{k}(P_n)}\left(P^0_{n,B_n}\right), \psi_0\right) + 2C(\lambda)\frac{1+\log(K(n))}{np},$$

where $\hat{k}(P_n) \equiv \text{argmin}_k E_{B_n} \int L\left(o, \hat{\Psi}_k\left(P^0_{n,B_n}\right)\right) dP_0(o)$ is the cross-validation selector, the comparable oracle selector is $\tilde{k}(P_n) \equiv \text{argmin}_k E_{B_n} \int L\left(o, \hat{\Psi}_k\left(P^0_{n,B_n}\right)\right) dP^1_{n,B_n}(o)$, and $P[\hat{\Psi}_k(P_n) \in \Psi] = 1$.

## 4. Experiments

The sample size and the number of covariates that have been used in Chernozhukov et al. (2018) were $N = 500, 1000$, with numbers of covariates chosen as of $p = 20$. Our simulation study has extended that by including $N = 100$, as relatively low sample size, and larger size of covariates, $p = (20, 50, 80, 100)$, $p = (200, 300, 500)$, and $p = (1000, 2000, 3000)$. Under these settings, two scenarios are introduced. Consider the true value of the average treatment effect is set to be $\beta_0 = 0.5$ in the SCM model

$$Y = T\beta_0 + g_0(\mathbf{X}) + U, \mathbb{E}[U \mid \mathbf{X}, D] = 0,$$

$$T = m_0(\mathbf{X}) + V, \mathbb{E}[V \mid \mathbf{X}] = 0.$$

And the nuisance parameters are (Bach et al., 2021)

$$g_0(x_i) = \frac{\exp(x_i)}{1+\exp(x_i)} + \frac{1}{4}x_i,$$

$$m_0(x_i) = x_i + \frac{1}{4}\frac{\exp(x_i)}{1+\exp(x_i)}.$$

**Scenario 1**

The following scenario is from Chernozhukov et al. (2018) and Bach et al. (2021). The error terms are

$$u_i \sim \mathcal{N}(0,1),$$

$$v_i \sim \mathcal{N}(0,1),$$

with the covariates

$$x_i \sim \mathcal{N}(0,\Sigma), \ \Sigma_{kj} = 0.7^{|j-k|}.$$

**Scenario 2**

Consider the simulation studies introduced in Chernozhukov et al. (2018), Farbmacher et al. (2020), and Bach et al. (2021),

$$x_i \sim \mathcal{N}(0,\Sigma), \ \Sigma_{kj} = 0.5^{|j-k|},$$

and

$$\begin{pmatrix} U \\ V \end{pmatrix} \sim \mathcal{N}\left(0, \begin{pmatrix} 1 & 0.3 \\ 0.3 & 1 \end{pmatrix}\right).$$

Table 1 summarizes the planned simulation scheme with the scenarios and the cases.

**Table 1**
*Simulation Scheme Plan*

| Levels of High Dimensional Data | Low-High-Dimensional (LHD) | | | | | | | | | Moderate-High-Dimensional (MHD) | | | | | | | | | Big-High-Dimensional (BHD) | | | | | | | | | |
|---|---|---|---|---|---|---|---|---|---|---|---|---|---|---|---|---|---|---|---|---|---|---|---|---|---|---|---|
| *Number of* Covariates (*p*) | 20 | | | 50 | | | 80 | | | 100 | | | 200 | | | 500 | | | 1000 | | | 2000 | | | 5000 | | |
| *Sample* Size (*N*) | 100 | 500 | 1000 | 100 | 500 | 1000 | 100 | 500 | 1000 | 100 | 500 | 1000 | 100 | 500 | 1000 | 100 | 500 | 1000 | 100 | 500 | 1000 | 100 | 500 | 1000 | 100 | 500 | 1000 |
| Scenario 1 | Case1 | Case 2 | Case 3 | Case 4 | Case 5 | Case 6 | Case 7 | Case 8 | Case 9 | Case 10 | Case 11 | Case 12 | Case 13 | Case 14 | Case 15 | Case 16 | Case 17 | Case 18 | Case 19 | Case 20 | Case 21 | Case 22 | Case 23 | Case 24 | Case 25 | Case 26 | Case 27 |
| Scenario 2 | Case 28 | Case 29 | Case 30 | Case 31 | Case 32 | Case 33 | Case 34 | Case 35 | Case 36 | Case 37 | Case 38 | Case 39 | Case 40 | Case 41 | Case 42 | Case 43 | Case 44 | Case 45 | Case 46 | Case 47 | Case 48 | Case 49 | Case 50 | Case 51 | Case 52 | Case 53 | Case 54 |

# 5. Results

To evaluate the performance of the causal inference in double machine learning for the semiparametric approach, the two scenarios described in the simulation scheme are implemented to compare the three SCM models. The first scenario has a data generation process with uncorrelated covariance, while the second scenario allows a correlated covariance. For each scenario, three levels of the high dimensional data setting are adopted: low-high-dimensional (LHD) for $p = (20, 50, 80)$, moderate-high-dimensional (MHD) for $p = (100, 200, 500)$, and Big-high-dimensional (BHD) for $p = (1000, 2000, 5000)$.

The method is applied to the real data to explore new findings and compare performance. For each research question, there are two parts, the first addresses the simulation performance of the methods, and the second undertakes the application of the three methods to the real data.

In this simulation, there are 3 different sample sizes, 100, 500, and 1000. For each sample size, there are 9 categories of $p$ covariates, 20, 50, 80, 100, 200, 500, 1000, 2000, 5000. And two scenarios of correlated and uncorrelated errors. There is a total of 54 simulations for each of the 3 research questions, which make up 162 simulations. Those simulations are operated using both the personal computers and the high-performance computing cluster (HPC), which has assisted to make an informative performance comparison between the two computing paradigms and serves as a reference for future replication. The personal computers were operated with different cores ranging from 4-20 cores.

## Simulation Results of Research Question 1

Simulation Results of Research Question 1 concerning support vector machine (SVM) model through Scenario 1 are displayed under the three levels of covariates size: LHD, MHD, and BHD.

### *Simulation Results of Research Question 1 for Low-High-Dimensional Data*

**Table 2**
*Results of Question 1 for Scenario 1 with Low-High-Dimensional Data when p = (20, 50, 80)*

| Scenario 1 | $N$ | Bias | SE | SE-adjusted | MSE | Time |
|---|---|---|---|---|---|---|
|  | 100 | 0.0150 | 0.0882 | 0.0089 | 0.0080 | 4.8092 |
| $p = 20$ | 500 | 0.0158 | 0.0382 | 0.0018 | 0.0017 | 18.7706 |
|  | 1000 | -0.0004 | 0.0294 | 0.0009 | 0.0009 | 65.6771 |
|  | 100 | 0.2118 | 0.0657 | 0.0222 | 0.0492 | 5.5384 |
| $p = 50$ | 500 | 0.1901 | 0.0360 | 0.0087 | 0.0374 | 41.2833 |
|  | 1000 | 0.1283 | 0.0256 | 0.0041 | 0.0171 | 147.055 |
|  | 100 | 0.2210 | 0.0721 | 0.0233 | 0.0541 | 6.2527 |
| $p = 80$ | 500 | 0.1616 | 0.0349 | 0.0073 | 0.0273 | 66.0967 |
|  | 1000 | 0.2008 | 0.0232 | 0.0064 | 0.0409 | 223.6176 |

*Note.* The number of replications is 500, $N$ = sample sizes of (100, 500, 1000), Time = the running time of computing. These simulations were run with PC's.

## Simulation Results of Research Question 1 for Moderate-High-Dimensional Data

**Table 3**

*Results of Question 1 for Scenario 1 with Moderate-High-Dimensional Data when p = (100, 200, 500)*

| Scenario 1 | $N$ | Bias | SE | SE-adjusted | MSE | Time |
|---|---|---|---|---|---|---|
|  | 100 | 0.2049 | 0.0706 | 0.0217 | 0.0470 | 10.4183 |
| $p = 100$ | 500 | 0.1495 | 0.0325 | 0.0068 | 0.0234 | 22.916 |
|  | 1000 | 0.1785 | 0.0234 | 0.0057 | 0.0324 | 55.3211 |
|  | 100 | 0.1818 | 0.0743 | 0.0196 | 0.0386 | 7.7703 |
| $p = 200$ | 500 | 0.1343 | 0.0342 | 0.0062 | 0.01920 | 23.7825 |
|  | 1000 | 0.1976 | 0.0204 | 0.0062 | 0.0394 | 86.0439 |
|  | 100 | 0.2284 | 0.0729 | 0.0240 | 0.0575 | 13.8357 |
| $p = 500$ | 500 | 0.1467 | 0.0309 | 0.0067 | 0.0224 | 42.6681 |
|  | 1000 | 0.1673 | 0.0221 | 0.0053 | 0.0285 | 130.3802 |

*Note.* The number of replications is 500, $N$ = sample sizes of (100, 500, 1000), Time = the running time of computing. These simulations were run with the high-performance computing (HPC).

## Simulation Results of Research Question 1 for Big-High-Dimensional Data

**Table 4**

*Results of Question 1 for Scenario 1 with Big-High-Dimensional Data when p = (1000, 2000, 5000)*

| Scenario 1 | $N$ | Bias | SE | SE-adjusted | MSE | Time |
|---|---|---|---|---|---|---|
|  | 100 | 0.0965 | 0.0573 | 0.0112 | 0.0126 | 15.1813 |
| $p = 1000$ | 500 | 0.1615 | 0.0318 | 0.0074 | 0.0271 | 82.3944 |
|  | 1000 | 0.1246 | 0.0235 | 0.0040 | 0.0161 | 257.2423 |
|  | 100 | 0.1405 | 0.0663 | 0.0155 | 0.0241 | 49.631 |
| $p = 2000$ | 500 | 0.1800 | 0.0315 | 0.0082 | 0.0334 | 158.3181 |
|  | 1000 | 0.1894 | 0.0229 | 0.0060 | 0.0364 | 568.4668 |
|  | 100 | 0.1962 | 0.0705 | 0.0209 | 0.0435 | 140.748 |
| $p = 5000$ | 500 | 0.1960 | 0.0331 | 0.0089 | 0.0395 | 505.7694 |
|  | 1000 | 0.1764 | 0.0252 | 0.0056 | 0.0318 | 1579.582 |

*Note.* The number of replications is 500, $N$ = sample sizes of (100, 500, 1000), Time = the running time of computing. The simulations were run with the high-performance computing (HPC).

# Simulation Results Of Research Question 2

The results of Research Question 2 concerning the DL through Scenario 1 are displayed under the three levels of covariates size: LHD, MHD, and BHD.

## *Simulation Results Of Research Question 2 for Low-High- Dimensional Data*

**Table 5**
*Results of Question 2 for Scenario 1 with Low-High-Dimensional Data when p = (20, 50, 80)*

| Scenario 1 | N | Bias | SE | SE-adjusted | MSE | Time |
|---|---|---|---|---|---|---|
| | 100 | 0.1769 | 0.0779 | 0.0193 | 0.0374 | 1.5285 |
| p = 20 | 500 | 0.1730 | 0.0350 | 0.0079 | 0.0311 | 40.1626 |
| | 1000 | 0.1694 | 0.0247 | 0.0054 | 0.0293 | 280.8999 |
| | 100 | 0.1716 | 0.0774 | 0.0188 | 0.0354 | 2.3883 |
| p = 50 | 500 | 0.1704 | 0.0352 | 0.0078 | 0.0303 | 37.8766 |
| | 1000 | 0.1722 | 0.0247 | 0.0055 | 0.0303 | 125.0593 |
| | 100 | 0.1795 | 0.0782 | 0.0196 | 0.0383 | 6.0845 |
| p = 80 | 500 | 0.1755 | 0.0351 | 0.0080 | 0.0320 | 178.9186 |
| | 1000 | 0.1702 | 0.0247 | 0.0054 | 0.0296 | 207.3817 |

*Note.* The number of replications is 500, $N$ = sample sizes of (100, 500, 1000), Time = the running time of computing. These simulations were run with PC's.

## *Simulation Results of Research Question 2 for Moderate-High-Dimensional Data*

**Table 6**
*Results of Question 2 for Scenario 1 with Moderate-High-Dimensional Data when p = (100, 200, 500)*

| Scenario 1 | N | Bias | SE | SE-adjusted | MSE | Time |
|---|---|---|---|---|---|---|
| | 100 | 0.1731 | 0.0775 | 0.0187 | 0.036 | 4.1952 |
| p =100 | 500 | 0.173 | 0.0349 | 0.0079 | 0.0311 | 8.7394 |
| | 1000 | 0.1701 | 0.0248 | 0.0054 | 0.0296 | 24.4785 |
| | 100 | 0.1771 | 0.0783 | 0.0192 | 0.0375 | 3.1304 |
| p =200 | 500 | 0.1692 | 0.0349 | 0.0077 | 0.0299 | 15.0348 |
| | 1000 | 0.1712 | 0.0247 | 0.0055 | 0.0299 | 43.3119 |
| | 100 | 0.1762 | 0.0791 | 0.0191 | 0.0373 | 6.8924 |
| p =500 | 500 | 0.1719 | 0.0351 | 0.0078 | 0.0308 | 35.6161 |
| | 1000 | 0.1703 | 0.0248 | 0.0054 | 0.0296 | 109.1765 |

*Note.* The number of replications is 500, $N$ = sample sizes of (100, 500, 1000), Time is the running time of computing. These simulations were run with the high-performance computing (HPC).

## Simulation Results Question 2 for Big- High- Dimensional Data

**Table 7**
*Results of Question 2 for Scenario 1 with Big-High-Dimensional Data when p = (1000, 2000, 5000)*

| Scenario 1 | N | Bias | SE | SE-adjusted | MSE | Time |
|---|---|---|---|---|---|---|
|  | 100 | 0.1867 | 0.0782 | 0.0204 | 0.041 | 10.7405 |
| p = 1000 | 500 | 0.1763 | 0.0352 | 0.0080 | 0.0323 | 67.0095 |
|  | 1000 | 0.1713 | 0.0248 | 0.0055 | 0.0300 | 221.0456 |
|  | 100 | 0.1782 | 0.0783 | 0.0194 | 0.0379 | 23.6871 |
| p = 2000 | 500 | 0.1747 | 0.0352 | 0.0080 | 0.0318 | 139.5206 |
|  | 1000 | 0.1693 | 0.0247 | 0.0054 | 0.0293 | 500.3704 |
|  | 100 | 0.1792 | 0.0779 | 0.0195 | 0.0382 | 94.3971 |
| p = 5000 | 500 | 0.1731 | 0.0349 | 0.0079 | 0.0312 | 493.4778 |
|  | 1000 | 0.1718 | 0.0248 | 0.0055 | 0.0301 | 1430.124 |

*Note.* The number of replications is 500, $N$ = sample sizes of (100, 500, 1000), Time = the running time of computing. These simulations were run with the high-performance computing (HPC).

# Simulation Results of Research Question 3

The results of Research Question 3 concerning the hybrid method (SDL) through Scenario 1 are displayed under the three levels: LHD, MHD, and BHD.

## Simulation Results of Question 3 for Low-High-Dimensional Data

**Table 8**
*Results of Question 3 for Scenario 1 with Low-High-Dimensional Data when p = (20, 50, 80)*

| Scenario 1 | N | Bias | SE | SE-adjusted | MSE | Time |
|---|---|---|---|---|---|---|
|  | 100 | 0.029 | 0.0751 | 0.0097 | 0.0065 | 82.8915 |
| p = 20 | 500 | -0.0179 | 0.0338 | 0.0019 | 0.0015 | 117.4088 |
|  | 1000 | -0.0249 | 0.0240 | 0.0011 | 0.0012 | 234.6404 |
|  | 100 | 0.0547 | 0.0749 | 0.0105 | 0.0086 | 78.4494 |
| p = 50 | 500 | -0.0094 | 0.0337 | 0.0019 | 0.0012 | 172.9981 |
|  | 1000 | -0.0189 | 0.0239 | 0.001 | 0.0009 | 453.1429 |
|  | 100 | 0.0529 | 0.0759 | 0.0102 | 0.0086 | 85.5482 |
| p = 80 | 500 | -0.004 | 0.0338 | 0.0019 | 0.0012 | 1872.9109 |
|  | 1000 | -0.016 | 0.0239 | 0.0010 | 0.0008 | 677.2905 |

*Note.* The number of replications is 500, $N$ = sample sizes of (100, 500, 1000), Time = the running time of computing. These simulations were run with PCs.

*Simulation Results Question 3 for Moderate-High-Dimensional Data*

**Table 9**

*Results of Question 3 for Scenario 1 with Moderate-High-Dimensional Data when p = (100, 200, 500)*

| Scenario 1 | $N$ | Bias | SE | SE-adjusted | MSE | Time |
|---|---|---|---|---|---|---|
| | 100 | 0.0667 | 0.0743 | 0.0108 | 0.0100 | 93.6581 |
| $p = 20$ | 500 | -0.0053 | 0.0336 | 0.0018 | 0.0012 | 120.6646 |
| | 1000 | -0.0152 | 0.0239 | 0.0010 | 0.0008 | 136.0735 |
| | 100 | 0.0758 | 0.0749 | 0.0113 | 0.0114 | 80.1428 |
| $p = 50$ | 500 | -0.0011 | 0.0335 | 0.0018 | 0.0011 | 130.3356 |
| | 1000 | -0.0133 | 0.0238 | 0.0010 | 0.0007 | 244.8391 |
| | 100 | 0.0843 | 0.0761 | 0.0120 | 0.0129 | 85.8635 |
| $p = 80$ | 500 | 0.0043 | 0.0334 | 0.0019 | 0.0011 | 175.2444 |
| | 1000 | -0.0107 | 0.0238 | 0.0009 | 0.0007 | 241.7942 |

*Note.* The number of replications is 500, $N$ = sample sizes of (100, 500, 1000), Time = the running time of computing. These simulations were run with the high-performance computing (HPC).

*Simulation Results Question 3 for Big-High-Dimensional Data*

**Table 10**

*Results of Question 3 for Scenario 1 with Big-High-Dimensional Data when p = (1000, 2000, 5000)*

| Scenario 1 | $N$ | Bias | SE | SE-adjusted | MSE | Time |
|---|---|---|---|---|---|---|
| | 100 | 0.1088 | 0.0773 | 0.0135 | 0.0178 | 99.0878 |
| $p = 1000$ | 500 | 0.0058 | 0.0333 | 0.0018 | 0.0011 | 188.4077 |
| | 1000 | -0.007 | 0.0237 | 0.0009 | 0.0006 | 385.9213 |
| | 100 | 0.1109 | 0.0762 | 0.0136 | 0.0181 | 141.0441 |
| $p = 2000$ | 500 | 0.0127 | 0.0333 | 0.0019 | 0.0013 | 304.3408 |
| | 1000 | -0.0038 | 0.0237 | 0.0009 | 0.0006 | 704.1351 |
| | 100 | 0.1225 | 0.0772 | 0.0146 | 0.0210 | 402.0184 |
| $p = 5000$ | 500 | 0.0135 | 0.0332 | 0.0018 | 0.0013 | 915.3463 |
| | 1000 | 0.0015 | 0.0237 | 0.0009 | 0.0006 | 1984.354 |

*Note.* The number of replications is 500, $N$ = sample sizes of (100, 500, 1000), Time = the running time of computing. These simulations were run with the high-performance computing (HPC).

## Summaries of the Simulation Study

The following summaries compare the mean square error value, and the computational time of simulations cross the three models: SVM, DL, and the hybrid SDL. Under the three levels of high dimensional data: LHD, MHD and BHD.

Table 11 shows that the best MSE was achieved by SDL method with MSE = 0.0006 for BHD, MSE = 0.0007 for MHD, and MSE = 0.0008 for LHD, followed by SVM method with MSE = 0.009 for LHD.

**Table 11**
*Mean Square Error for the Three Methods (Support Vector Machine, Deep Learning, and Super Deep Learning) under the Three Level of Data in Scenario 1 and Scenario 2*

|     | High Dimensional Data Levels | Scenarios | SVM | DL | SDL |
| --- | --- | --- | --- | --- | --- |
| MSE | Low-High-Dimensional (LHD) | S1 | 0.0009 | 0.0293 | 0.0008 |
|     |                            | S2 | 0.072  | 0.0844 | 0.0136 |
|     | Moderate-High-Dimensional (MHD) | S1 | 0.0192 | 0.0296 | 0.0007 |
|     |                                 | S2 | 0.0511 | 0.0839 | 0.0150 |
|     | Big-High-Dimensional (BHD) | S1 | 0.0126 | 0.0293 | 0.0006 |
|     |                            | S2 | 0.0321 | 0.0841 | 0.0176 |

*Note*. DL = deep learning method, SDL = the hybrid of super learner and deep learning method, SVM = support vector machine method, Time = the computer running time for simulation.

Table 12 shows that the lowest total computational time was for DL method in scenario 1 for MHD and BHD data, and in Scenario 2 for LHD.

**Table 12**
*Time of Computation for the Three Methods (Support Vector Machine, Deep Learning, and the Hybrid Super Deep Learning) Under the Three Data Levels Sor Scenario 1 and Scenario 2*

|                | High Dimensional Data Levels | Scenarios | SVM | DL | SDL |
| --- | --- | --- | --- | --- | --- |
| Time (Minutes) | Low-High-Dimensional (LHD) | S1 | 579.1006 | 880.3000 | 3775.281 |
|                |                            | S2 | 904.7537 | 882.3153 | 2183.793 |
|                | Moderate-High-Dimensional (MHD) | S1 | 393.1361 | 250.5752 | 1308.616 |
|                |                                 | S2 | 419.7998 | 263.9633 | 1329.160 |
|                | Big-High-Dimensional (BHD) | S1 | 3342.152 | 2969.632 | 5025.568 |
|                |                            | S2 | 3417.592 | 3039.296 | 5217.317 |

*Note*. DL = deep learning method, SDL = the hybrid of super learner and deep learning method, SVM = support vector machine method, Time = the running time for simulation.

## Real Data Analysis

We revisit the real data 401(k) plan used in Chernozhukov et al. (2018) with the goal to estimate the effect of 401(k) eligibility on the total financial assets. The treatment T is defined as 401(k) eligibility, and the response is the total financial assets. The covariates vector X consists of 9 variables, age of participants, income, family size, years of education, individual defined benefit pension, marital status, individual participation in IRA plan, homeowner, and two-earner household.

The data were analyzed using the DML with SPSS methods, including SVM, DL, and the hybrid SDL method. A comparison with the literature (Chernozhukov et al., 2018) uses Lasso with k-fold sample splitting. In the data analysis, variable normalization was applied to simplify the results and accommodating the support points sample splitting requirement.

Table 13 shows the results of the real data analysis on 401 (k) plan, comparing outcomes after variable normalization with those obtained using Lasso DML with k-fold sample splitting. Our DML approach uses SVM, DL, and SDL with SPSS. The hybrid DML method of SDL with SPSS displayed the best computational efficiency with a time of 0.0429 seconds. followed by DL with 1.1610 seconds, where both methods outperformed those of Chernozhukov et al. (2018) using Lasso and k-fold sample splitting. The DML method using SVM with SPSS provided the most accurate estimation with $SE= 0.0006$, followed by the DML method using DL with SPSS ($SE= 0.0056$), and then the hybrid SDL ($SE= 0.0065$). The method using Lasso with k-fold sample splitting showed the lowest estimation performance, with $SE= 0.0071$.

The results of the real data study are highly consistent with those of the simulation study. Classic methods like SVM demonstrated good performance with Low-High-Dimensional data in the simulation study; however, in terms of the computational time, they do not outperform DL and SDL. The reason we see results such as SDL performing better than DL in computational time may be due to the 401(k) plan data having a relatively small number of covariates ($p = 11$), a range not covered in the simulation study, where the number of covariates in Low- High- dimensional was p= (20, 50, 80). Additionally, the classic Lasso method aligns closely with the performance and findings of the SVM method.

**Table 13**
*Comparison of Real Data DML Analysis Among the Methods*

|  | **Lasso-K Fold** | **SVM-SPSS** | **DL-SPSS** | **SDL-SPSS** |
|---|---|---|---|---|
| Estimator | 0.0030 | 0.0056 | 0.0095 | 0.0063 |
| *SE* | 0.0071 | 0.0006 | 0.0056 | 0.0065 |
| Time (Seconds) | 3.4870 | 28.7207 | 1.1610 | 0.0429 |

*Note*. DL-SPSS= the DML of deep learning (DL) with support point sample splitting, SDL-SPSS = the hybrid DML method of super learner and deep learning with support point sample splitting, SVM-SPSS = the DML of support vector machine (SVM) with support points splitting method.

# 6. Conclusion and Discussion

This study contributed to the literature by providing insight into the performance of the three DML methods for causal inference using SPSS. The best-performing method, with the lowest MSE, was the hybrid method SDL using SPSS under the proposed settings.

However, the DML of DL with SPSS demonstrated the best simulation time efficiency across both data scenarios and for three levels of high dimensional data: the low-high-dimensional (LHD), the moderate-high-dimensional (MHD), and in big-high-dimensional (BHD). The DML of SVM with SPSS did not perform well in high dimensional data settings, either in estimation accuracy or computational time compared to the other methods. While SVM showed good estimation performance within the low dimensional data framework for Scenario 1, did not outperformed the hybrid DML of SDL in the simulation study under moderate-high-dimensional (MHD) and big-high-dimensional (LHD).

This study suggests that the DML of SDL is recommended when estimation quality is prioritized. However, if time efficiency is preferred, the deep learning method with SPSS is the best option.The study does not recommend relying solely on traditional machine learning methods like SVM, as advanced methods such as SL and DL demonstrated superior performance in both estimation accuracy and computational efficiency.

Machine learning algorithms require high performance computing resources. To address these demands, we utilized the Rocky Mountain Advanced Computing Consortium (RMACC) provided by the University of Colorado at Boulder. Given the computational limitations encountered, future work could explore ways to make high-end hardware more accessible for common use or to develop less computationally intensive machine learning algorithms, possibly through specialized computing machines tailored to manage these tasks.

Causal inference from observational data is gaining increasing popularity recently, as it provides insights into cause-effect relationships rather than solely focusing on prediction. Advances in this area will contribute significantly not only to the statistics field but also to the applied sciences, including health sciences, social sciences, and economics in observational studies.

## Acknowledgement

The authors would like to extend their acknowledgments and thanks to C. Hansen for sharing the work of double machine learning from his paper Chernozhukov et al. (2018). And Thanks to K. Varaku for sharing her work on double machine learning.